\documentclass[11pt]{article}
\usepackage[a4paper, margin=1in]{geometry}
\usepackage{amsmath,amssymb} 
\usepackage[utf8]{inputenc} 
\usepackage{hyperref}       
\usepackage{url}            
\usepackage{booktabs}       
\usepackage{microtype}      
\usepackage[ ruled]{algorithm2e}
\usepackage{algorithmic,  array}
\usepackage{graphicx}
\newcommand{\argmin}{\arg\min}
\usepackage{color}

\DeclareUnicodeCharacter{B0}{\textdegree}
\title{Training Skinny Deep Neural Networks with Iterative Hard Thresholding Methods}

\author{Xiaojie Jin$^1$ Xiaotong Yuan$^4$ Jiashi Feng$^2$ Shuicheng Yan$^{3,2}$\\
\small $^1$NUS Graduate School for Integrative Science and Engineering, NUS\\
\small$^2$Department of ECE, NUS \qquad $^3$360 AI Institute\\
\small $^4$School of Information and Control, Nanjing University of Information Science \& Technology\\
\tt\small xiaojie.jin@nus.edu.sg, xtyuan1980@gmail.com,  \\
\tt\small elefjia@nus.edu.sg, yanshuicheng@360.cn}
\begin{document}
\date{}
\maketitle

\begin{abstract}
  Deep neural networks have achieved remarkable success in a wide range of
  practical problems. However, due to the inherent large parameter space, deep models are
  notoriously prone to overfitting and
  difficult to be deployed in portable devices with limited memory. In this
  paper, we propose an iterative hard thresholding (IHT) approach to train  Skinny Deep Neural Networks (SDNNs). An SDNN has much fewer parameters  yet can achieve competitive or even better performance than its full CNN counterpart. More concretely, the IHT approach trains an SDNN through following two alternative phases: (I) perform hard thresholding to drop connections with small activations and fine-tune the other significant filters; (II)~re-activate the frozen connections and train the entire network to improve its overall
  discriminative capability. We verify the superiority of SDNNs in terms of efficiency and classification performance 
   on four benchmark object recognition datasets, including CIFAR-10, CIFAR-100, MNIST and
  ImageNet. Experimental results clearly demonstrate that IHT can be applied for training SDNN based on various CNN architectures such as NIN
  and AlexNet. 
\end{abstract}

\section{Introduction}
Deep neural networks (DNNs) have achieved remarkable success in various applications. This has been driven by the rapid growth in the size of datasets, increasingly deeper network architectures
and the development of various techniques in training deep models. 

Despite their strong
capability of learning rich and discriminative representations, DNNs usually suffer from following two
problems caused by their inherent huge parameter space when applied in practice. First, most of state-of-the-art DNN architectures  are prone to
over-fitting even trained on large datasets~\cite{vgg,googlenet}. Secondly,
as they usually consume large storage memory and  computational resource, it is difficult to
embed modern DNN models into devices with limited power and memory, \emph{e.g.}, mobile phones.

A lot of regularization techniques have thus been proposed to decrease the risk of over-fitting for DNNs, \emph{e.g.}, dropout~\cite{dropout}. However, those techniques are unable to reduce the
storage cost. Recently, several works have been devoted to compressing and
speeding-up DNNs by pruning internal layer connections. However, most
of those methods achieve efficiency gain at the cost of performance deterioration.

In this work, we propose a novel approach for training skinny DNNs (SDNNs) with explicit size constraints to address the above problems simultaneously,
\emph{i.e.}, reducing the risk of overfitting to improve the generalization capability of deep
models and meanwhile reducing the model size to decrease storage memory cost. Compared
with other sparsity pursuit methods for training DNNs~\cite{han2015learning,han2016}, SDNN is superior as it is able to boost the
performance even when the model compression rate is
high. Briefly, our proposed approach for training SDNNs contains two alternative phases:  
\begin{itemize}
	\item \textbf{Phase I} \quad {Hard thresholding
	over connections and  sub-network  fine-tuning}. We apply hard thresholding over connections (weight parameters) at each layer  to select the  most prominent ones for the DNN model. The hard thresholding 
	preserves the top $k$ weight parameters with the largest magnitude and disables the others by zeroing their values. Then, we fine-tune the non-disabled parameters for compensating
	the performance loss caused by reducing the number of filters.
	\item \textbf{Phase II}  \quad  {Connection restoration and training the entire network}. The frozen connections are re-activated and
	all the parameters  are learned through training the entire network. The goal of this phase is
	to restore the truncated parameters and involve them  in learning
	better representations.
\end{itemize}
Alternating the above two phases in training
DNNs is able to produce SDNNs which have a stronger generalization capability
with fewer parameters compared with the counterpart trained in the conventional
way. We term such a method compositing of the above two phases as iteratively hard thresholding (ITH).  Note that these two alternative  phases are only needed in  training SDNNs, while in the
testing stage SDNN only takes one feedforward pass for inputs to make prediction.

To verify the effectiveness of SDNNs and the ITH training approach, we conduct extensive experiments on four
 public datasets with various scales, \emph{i.e.}, MNIST~\cite{lecun1998gradient}, CIFAR10~\cite{krizhevsky2009learning}, CIFAR100~\cite{krizhevsky2009learning} and ImageNet~\cite{imagenet} for
two DNN architectures with different complexities including Network
in Network~\cite{nin}, and AlexNet~\cite{krizhevsky2012imagenet}. The experimental results clearly demonstrate that SDNN with ITH does not
only improve the generalization capability of deep models and provide 
state-of-the-art performance, but also reduce the size of parameters at the
same time. Therefore, ITH is a quite appealing approach for training SDNNs in real-world scenarios concerning limited computational and storage cost.

\section{Preliminaries: Gradient Hard Thresholding  Revisit}
\label{sec:ght}
The gradient hard thresholding  (GHT) algorithm was proposed by~\cite{yuan2013gradient} to solve the
following sparsity-constrained convex optimization problem:
\begin{equation}
  \mathop {\min }\limits_{\mathbf{x} \in \mathbb{R}^d } f(\mathbf{x}), \text{ s.t. } \|\mathbf{x}\|_0  \le k,
\label{eq:ght}
\end{equation}
where $f:\mathbb{R}^d\mapsto\mathbb{R}$ is a smooth convex function and
$\|\mathbf{x}\|_0$ counts  nonzero elements in $\mathbf{x}$.  The GHT algorithm solves the problem in
Eqn.~\eqref{eq:ght} by alternatively performing gradient descent and hard
thresholding over the gradient. More concretely, let $\mathcal{O}_k(\mathbf{x})$ denote the hard
thresholding operator which selects the top $k$  entries of
$\mathbf{x}$ with largest magnitudes and set the rest entries to 0. Let $\mathbf{x}^{(t)}$ denote the
updated variable and $F^{(t)}$ denote the support set of
$\mathcal{O}_k(\mathbf{x}^{(t)})$  at the $t$-th
iteration. We further define $\mbox{supp}(\mathbf{x})$ as the support set of
$\mathbf{x}$ and $\mbox{supp}(\mathbf{x},k)$ as the index set of the top $k$
entries of $\mathbf{x}$. At the $t$-th iteration, following three steps (denoted
as \emph{S1}, \emph{S2} and \emph{S3}, respectively) are involved in GHT:

\emph{S1}: perform gradient descent at $\mathbf{x}^{(t-1)}$ with a
step-size $\eta$:
$\tilde{\mathbf{x}}^{(t)} = \mathbf{x}^{(t - 1)} - \eta \nabla f(\mathbf{x}^{(t
  - 1)})$,
where $\nabla f(\mathbf{x}^{(t - 1)})$ is the gradient of $f(\cdot)$ evaluated at
$\mathbf{x}^{(t - 1)}$;

\emph{S2}: apply hard thresholding $\mathcal{O}_k(\cdot)$ on $\tilde{\mathbf{x}}^{(t)}$ as:
$\tilde{\mathbf{x}}^{(t)} = \mathcal{O}_k(
\tilde{\mathbf{x}}^{(t)})$.
Therefore, $F^{(t)} = \mbox{supp}(\tilde{\mathbf{x}},k)$;

\emph{S3}: optimize $\mathbf{x}^{(t)}$ by minimizing the objective
function over support set $F^{(t)}$, \emph{i.e.},
$\mathbf{x}^{(t)} = \mbox{arg min}\{f(\tilde{\mathbf{x}}),
\mbox{supp}(\mathbf{x}) \subseteq F^{(t)})\}$.

It is proved that under mild conditions, the GHT algorithm converges
geometrically to the point with bounded deviation from  global optimum, with a high probability~\cite{yuan2013gradient}.
\begin{algorithm}
\label{algorithm:ccodnn}
\SetKwInOut{Input}{input}\SetKwInOut{Output}{output}
\caption{Iterative Hard Thresholding for Training SDNNs}
 \Input{Training sample $(\mathbf{x}, y^*)$, randomly initialized weights
  $\mathbf{W}=(\mathbf{W}^{(1)}, \ldots, \mathbf{W}^{(L)})$ for an $L$-layer SDNN. Parameters $ s_1 $ and $ s_2 $. Loss function $ \mathcal{L} $.}
 (\textbf{Step 1}) Train the SDNN model for $s_1$ epochs, \emph{i.e.}, $\mathbf{W} =
\mathop {\argmin }_\mathbf{W}\mathcal{L}(\mathbf{x}, y^*, \mathbf{W}) $.\\
\While{maximum number of epochs is not reached}
{
	\vspace{2mm}
\emph{Phase I}:\\
(\textbf{Step 2})  Apply hard thresholding operation $\mathcal{O}_{k_\ell}(\mathbf{x})$ to
$\mathbf{W}^{(\ell)}$:
$\mathbf{W}^{(\ell)} = \mathcal{O}_{k_\ell}(\mathbf{W}^{(\ell)})$.\\
(\textbf{Step 3}) $F^{(\ell)} = \mbox{supp}(\mathbf{W}^{(\ell)},k_\ell)$, $\ell = 1, \ldots ,L$.\\
(\textbf{Step 4}) Train the SDNN for $s_2$ epochs:
$\mathbf{W} = \mathop {\argmin }_\mathbf{W}\{ \mathcal{L} (\mathbf{x}, y^*,
\mathbf{W}), \mbox{supp}(\mathbf{W}^{(\ell)}) \subseteq F^{(\ell)}\}$.\\
\vspace{2mm}
 \emph{Phase II}:\\
(\textbf{Step 5}) Restore the connections truncated at \textbf{Step 2}.\\
(\textbf{Step 6}) Train the deep model for $s_1$ epochs:
$\mathbf{W} =
\mathop {\argmin }_\mathbf{W}\{ \mathcal{L}  (\mathbf{x}, y^*, \mathbf{W}) \}$.
\vspace{2mm}
}
\Output{$\mathbf{W}$}
\end{algorithm}

\section{Skinny Deep Neural Networks}
\label{sec:ccodnn}
\subsection{Deep Neural Networks with Cardinality Constraint}
\label{sec:notations}
For expression conciseness, we only consider the case with a single training sample. The
formulation for training with multiple samples  can be derived similarly as samples
are independent and the loss function is decomposable over samples. We denote a training sample as $(\mathbf{x}, y^*)$ where
$\mathbf{x}\in \mathbb{R}^d$ denotes the raw input data and $y^* \in \{1,\ldots,C\}$ is its
ground truth category label. Here $C$ is the total number of categories. We consider a
DNN model consisting of $L$ layers, each of which outputs a feature map, denoted
as ${\mathbf{X}}^{(\ell)}$ for layer $\ell$. Here ${\mathbf{X}}^{(0)}$ and ${\mathbf{X}}^{(L)}$
represent the input and final output of the network,
respectively. Let $\mathbf{W}^{(\ell)}$ denote  parameters of the filters (or weights)
of the $\ell$-th layer to be learned, and 
$\mathbf{W}=(\mathbf{W}^{(1)}, \ldots, \mathbf{W}^{(L)})$ is the collection of
all learnable parameters in a DNN. We use
$\mathbf{W}^*=(\mathbf{W}^{(1)^*}, \ldots, \mathbf{W}^{(L)^*})$ to denote
output the parameters  after training. 
Using the above notations, 
the output of each layer in an $ L $-layer DNN can be written as
\begin{equation*}
\label{activation}
\mathbf{X}^{(\ell)} = g^{(\ell)}(\mathbf{W}^{(\ell)}  * \mathbf{X}^{(\ell - 1)} ),\ \ \ell = 1, \ldots ,L\, \ \ \mbox{and} \ \ \mathbf{X}^{(0)} \triangleq \mathbf{x},
\end{equation*}
where $g^{(\ell)}( \cdot )$ is a composite of multiple
specific functions including activation function, dropout,
pooling, batch normalization and softmax. For succinct notations, the bias
term is omitted. The loss function of a deep model that we consider here is
\begin{equation}
\label{eq:2}
 \mathcal{L}(\mathbf{x}, y^*, \mathbf{W}) =
  -\log (h_{y^* } (\mathbf{x},\mathbf{W})) + \lambda \|\mathbf{W}\|_F,
\end{equation}
where $h_{y^* } (\mathbf{x},\mathbf{W})$ denotes the 
probability score predicted for $\mathbf{x}$ on the ground truth category of $y^*$, and $\lambda$ is the
weight decay factor. In traditional methods, a deep model is optimized through minimizing the loss function without any constraint over $\mathbf{W}$:
\begin{equation}
\label{eq:opt-tradition}
\min_\mathbf{W} \mathcal{L}(\mathbf{x}, y^*, \mathbf{W}).
\end{equation}
In this work, we propose to impose explicit cardinality constraints to layer-wise parameters during training in order to reduce the parameter size. Thus the optimization problem for training an SDNN is formulated as
\begin{equation}
\label{eq:opt-ccodnn}
\mathop {\min }\limits_\mathbf{W} \mathcal{L}(\mathbf{x}, y^*, \mathbf{W}), \text{ s.t. }
\|\mathbf{W}^{(\ell)^*}\|_0 \leq k_{\ell},\ \ \ell = 1, \ldots ,L,
\end{equation}
where $k_\ell$ is the cardinality constraint  for the parameters of the
$\ell$-th layer. Note that $k_{\ell}$ can be either the same or different across
different layers. Correspondingly, $r^{(\ell)}$ is denoted as the sparsity ratio of the $\ell$-th layer and defined as 
$r^{(\ell)} \triangleq \|\mathbf{W}^{(\ell)}\|_0/|\mathbf{W}^{(\ell)}|$ where $|\mathbf{W}^{(\ell)}|$ denotes the number of parameters at the $\ell$-th layer. 

\subsection{From GHT to SDNN}
\label{sec:from}
Inspired by the success of GHT on solving sparsity-constrained convex problems~\cite{yuan2013gradient},
we apply it to train DNNs to alleviate the over-parameterization
issue. However, straightforwardly applying GHT to train SDNNs cannot provide desirable results since the loss
 function is  non-convex  and highly complex. Therefore applying hard thresholding operation to the parameters of
deep model at each iteration in the same way as GHT will cause the model to diverge and be unable to learn
meaningful parameters.

In this paper, we propose a novel  approach for training SDNNs with cardinality
constraints. A summary of details on training SDNNs is
presented in Algorithm~\ref{algorithm:ccodnn}. Compared with GHT, there are two main differences in
the approach used for training SDNNs:

	 \textbf{Multi-Step Update}. \quad GHT performs hard thresholding operation at each
	iteration. However in training  SDNNs, we update the parameters for $s_1$ iterations before
	performing hard thresholding. Such a strategy yields following two advantages. First, the
	training is accelerated by largely reducing the time of performing hard
	thresholding. Secondly, by updating all parameters (including those of
	connections truncated by the last hard thresholding operation) sufficiently, the
	SDNNs are likely to learn more discriminative representation with more
	parameters. Otherwise with a single-step update, it is highly possible that
	connections which are truncated by the first hard thresholding operation would
	be always truncated in all subsequent hard thresholding operations due to the little
	change in magnitudes.
	
	 \textbf{Relaxation on Sparsity Constraints}. \quad  As can be seen from
	Eqn.~\eqref{eq:ght}, GHT applies the $\ell_0$ sparsity constraints to the parameters at
	each iteration. In contrast, since we only care about the final model after training
	is finished,  we do not need to optimize the SDNNs with sparsity
	constraints throughout the training stage. Actually, in SDNNs, we use such relaxation
	to restore the connections truncated by hard thresholding operation, aiming to
	learn better representations. More details are given in Section~\ref{sec:train-ccodnn}. As confirmed  by  experimental results, it is critical for
	SDNNs to learn more discriminative features than other network pruning
	works~\cite{han2015learning,han2016,anwar2015structured}.

\subsection{Training  SDNNs}
\label{sec:train-ccodnn}
The training of SDNN mainly consists of two iteratively alternating phases, each
of which contains an operation for the parameters of deep models followed by an
optimization process which is mutually different. In this section, we explain the
 details of  Algorithm \ref{algorithm:ccodnn}.

\textbf{Network Initialization:} \quad This corresponds to Step 1. At the beginning of the
training stage, all the parameters of the SDNN are trained for $s_1$ epochs without
considering cardinality constraints. Therefore the optimization formulation in this step
is exactly the same as the one in Eqn.~\eqref{eq:opt-tradition}. The aim of this step is to provide a good
initialization for the following steps, which prevents the SDNN from
diverging or getting stuck in a bad local minimum.

\textbf{Phase I:} \quad This phase corresponds to Step 2, Step 3 and Step 4 in Algorithm~\ref{algorithm:ccodnn}, which perform hard
thresholding  and sub-network fine-tuning, respectively. At Step 3, we keep
the top $k_\ell$ parameters with largest magnitudes at the $\ell$-th layer and set the rest
parameters to 0. At Step 4, we fine-tune the parameters reserved at Step 3. This
step aims  to compensate the performance drop caused by erasing a part of the parameters. As demonstrated
in our experiments, reinitializing the reserved parameters leads to poor
performance. This phenomenon could be explained similar as the observations made in~\cite{yosinski2014transferable}: since
DNNs contain fragile co-adapted features, the gradient descent algorithm is
able to find a good solution when the network is {trained from the scratch}, but it may fail
after re-initializing some layers and retraining them. Moreover, starting to fine-tune the
network  with retained weights requires less computation because it is not necessary to {train the entire network.} 

\begin{figure}[!t] \centering
\centering
	  \begin{minipage}[l]{0.45\textwidth}
	  \centering
	      \includegraphics[width=2.3in]{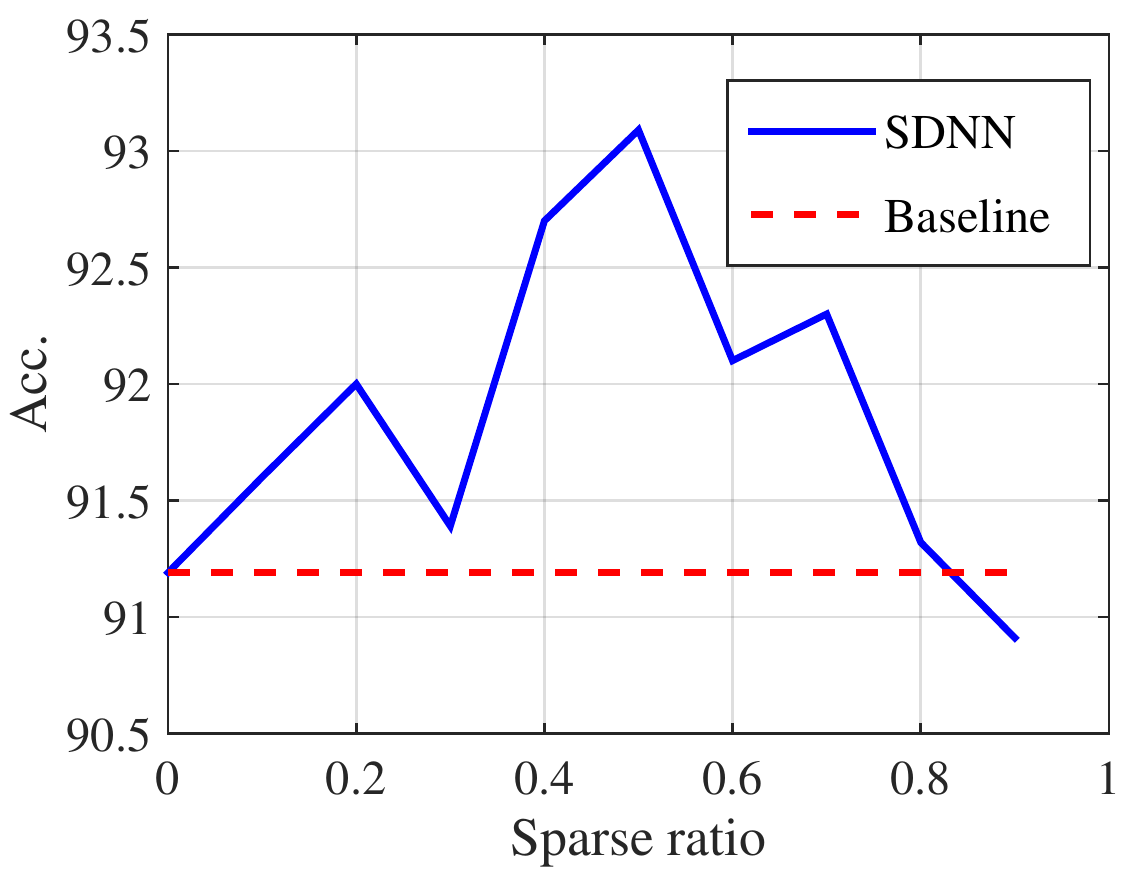}
	      \caption{The accuracy of SDNN on CIFAR10 with different sparse ratios. The baseline model is equal to a SDNN with sparse ratio $r = 0$. From $r=0.1$ to $r=0.8$, SDNN surpasses the baseline model while reducing the model size. Particular, when $r=0.5$, SDNN significantly improves the baseline by 1.9\%. When $r=0.9$, SDNN is outperformed by baseline model by only 0.28\%.}
	      \label{fig:sparse_ratio}
	  \end{minipage} \ \ \ 
    \begin{minipage}[r]{0.45\textwidth}
    \centering
        \includegraphics[width=2.3in]{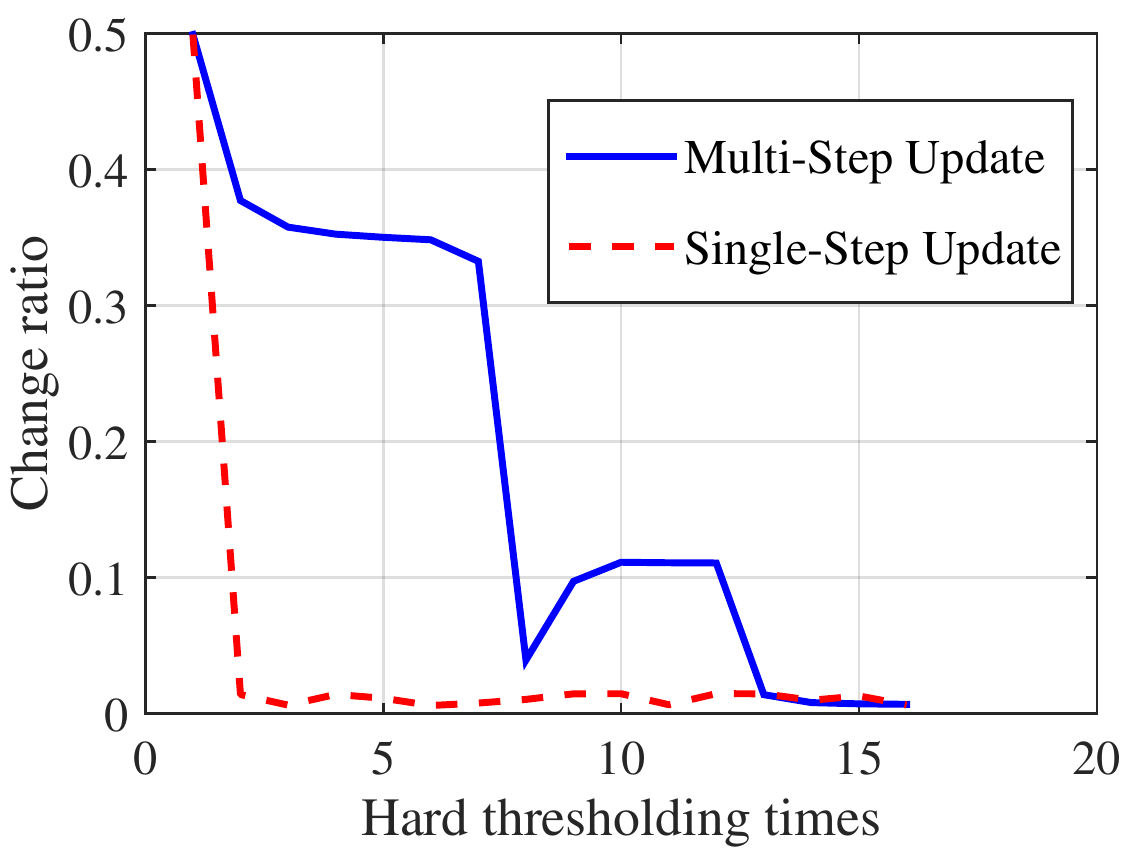}
        \caption{The change ratio of connection truncated by neighboring hard thresholding operations. The multi-step update adopted by SDNN has large change ratio in initial training phase, allowing the deep model to learn features with strong discriminative capability, while the single-step update would result $\sim$ 0.01\% change ratios during all training phase, thus undermines the representation capability of deep models.}
        \label{fig:ratio}

    \end{minipage}
	\vspace{-5.6mm}
\end{figure}

\textbf{Phase II:} \quad This phase corresponds to Step 5 and Step 6, which conduct a weight
restoration operation and trains the entire network, respectively. Step 5 removes
the cardinality constraints over parameters which are set to be 0 in Step 3 so that all
parameters are updated in Step 6 freely. This step is
critical in SDNN for the following reason: by updating all parameters including
those set to be 0 in Step 3, the SDNN is  able to restore some connections
that are beneficial for learning feature representation with strong discriminative
power. As can be seen in Figure~\ref{fig:ratio}, at the initial training stage, a large proportion of connections which are
truncated in the last round of hard thresholding operation become significant (with
parameters in large magnitudes) at this phase. Thus  SDNN has a
strong capability to search among a large parameter space for seeking a better local
optimum. Besides, we observe that the ratio decrease as the training progresses
due to the deep model converges to a good optimum.

\section{Related Works}
Early approaches for deep model compression including optimal brain
damage~\cite{lecun1989optimal} and optimal brain surgeon~\cite{hassibi1993second} that prune the connections in networks based
on  the second order information. However, those methods are not feasible for  deep networks due to high computational complexity. Recent
works aiming at network pruning include~\cite{han2015learning,collins2014memory,anwar2015structured,lebedev2015fast,kim2015compression}, which prune connections in a progressively greedy
way~\cite{han2015learning} or using sparsity related regularizer \cite{collins2014memory,lebedev2015fast}. Although those works
can reduce model size significantly, they suffer from the dramatic performance loss. In contrast, SDNN does not only offer significant compression
ratios but also improves the performance simultaneously.

Our work is also in line with model compression. For example, \cite{anwar2015structured} proposes to quantize the deep model by minimizing L2 error and \cite{denton2014exploiting} seeks 
an low-rank approximation of the model. Recently, \cite{han2016}
combined pruning, quantization and Huffman coding techniques and provided rather high compression
ratios. However, those methods also introduce performance drop. There are also works trying to compress a model by using binning
methods~\cite{gong2014compressing}, but they can only be applied over fully connected
layers. In contrast, our method can be applied for compressing both 
convolution layers and fully connected layers.

\section{Experiments and Analysis}
\label{sec:experiments}

\subsection{Experiment Settings and Implementation Details}
\label{sec:setting-detail}
We justify our method on four scale-various object classification benchmarks,
\emph{i.e.} three small-scale ones including CIFAR10~\cite{krizhevsky2009learning}, CIFAR100~\cite{krizhevsky2009learning} and MNIST~\cite{lecun1998gradient} and the
large-scale ImageNet~\cite{imagenet}. Two evaluation metrics, the classification performance and
the number of parameters in a model, are used in comparison with other methods.

\textbf{Deep Models}\quad In our experiments, we train and test two deep
models which are with different complexities, \emph{i.e.} Network in
Network (NIN)~\cite{nin}, and AlexNet~\cite{krizhevsky2012imagenet}. Briefly, NIN has only convolutional weight layers by replacing the single linear
convolution layers in the conventional CNNs by multilayer perceptrons and using
the global average pooling layer to generate feature maps for each
category. Compared with NIN, AlexNet is wider and deeper containing 60M
parameters with five convolution layers and three fully connected layers.


\textbf{Implementation}\quad All of our experiments are conducted on a NVIDIA
TITAN GPU using Caffe. The numbers of training epochs for Step 1 and Step 4 in
Algorithm 1 are set to $s_1=15$ and $s_2=150$ for small-scale datasets and
$s_1=15$ and $s_2=40$ for the ImageNet dataset to reduce training time. The
hyperparameters of NIN and AlexNet including learning rate, momentum and
weight decay follow~\cite{nin} and \cite{krizhevsky2012imagenet}, respectively. All the results of our
methods in the paper are based on the models reaching convergence. 

 Data augmentation is used in many models for object classification to prevent overfitting. The horizontal flipping is used for CIFAR10 and CIFAR100. For ImageNet, we use random crop and horizontal flipping as in~\cite{krizhevsky2012imagenet}. No data augmentation method is applied for MNIST. 

{
\textbf{Memory Usage}\quad  To efficiently utilize the sparsity property of SDNN to reduce the memory storage size, we refer to the ``Bitmask'' storage format proposed in~\cite{collins2014memory}, which stores the nonzero parameters as well as a mask whose number of bits is equal to the number of total parameters. The bit value will be set to 1 if the corresponding parameter is nonzero, otherwise to 0. As indicated by~\cite{collins2014memory}, such memory usage methods can be directly used in deep models at runtime and reduce the storage size efficiently at the same time.
}

\subsection{Model Analysis}
\label{sec:model-analysis}
In the following, we use NIN trained and tested on CIFAR10 to investigate the effects of
different 
$k_{\ell},\ \ \ell = 1, \ldots ,L$ (recall that $k_{\ell}$ denotes the number of nonzero parameters in the $\ell$-th layer) on the classification performance of SDNN as well as justify the hard thresholding strategy by comparing with the stochastic thresholding strategy. As a baseline, the NIN trained without sparsity constraint achieved 91.9\%~\cite{nin} with data augmentation on this dataset. 

\paragraph{Global Sparsity Distribution}\quad One problem in practical use is how to set the sparsity constraint $k_{(\ell)}$ for different layers. To ease the laborious work of tuning such parameters, especially when the number of layers in a deep model is large, \emph{e.g.} ResNet~\cite{residual}, we explore a simple way by setting all layers in a deep model with the same sparsity constraints. {As we have observed in the experiment, the training process of the network may diverge if directly applying hard thresholding to the layer-wise parameters when $r^{(\ell)}$ is large, \emph{e.g.}, 0.8. We propose a progressive hard thresholding strategy to address this issue, \emph{i.e.}, $r^{(\ell)}(t) = r^{(\ell)}(0) + t(r^{(\ell)}(T)-r^{(\ell)}(0))/T$ where $T$ is the overall training epochs, $r^{(\ell)}(t)$ denotes the sparsity ratio to layer-wise parameters when applying hard thresholding operation at $t$-th epochs. $r^{(\ell)}(0)$ and $r^{(\ell)}(T)$ are the sparsity ratio when the first hard thresholding operation is applied and the target layer-wise sparsity ratio, respectively.} The experimental results prove that such a simple strategy is able to boost the performance of the baseline model and simultaneously reduce the model size. Note although Collins \emph{et al.}~\cite{collins2014memory} proposed an iterative method to search the layer-wise sparsity hyperparameters in a greedy way, their method is time consuming thus not feasible for large deep models. In the experiments, we set the same sparsity ratio $r=\{0,0.1,0.2,\ldots,0.9\}$ for all layers. Note when $r=0$, our model degenerates to the baseline model. Figure \ref{fig:sparse_ratio} illustrates the curve of classification performance versus the sparsity ratio. 
As obviously observed in Figure \ref{fig:sparse_ratio}, SDNN outperforms the baseline model with the sparsity ratio ranging from 0.1 to 0.8, which demonstrates the strong capability of SDNN in enhancing the generalization capability of deep models. Particularly, when $r=0.5$, SDNN significantly increases the classification accuracy by 1.90\%. Even when the largest ratio is increased to $r=0.9$, \emph{i.e.} reducing the model size by 10$\times$, SDNN only bears a slight performance loss (0.28\%) compared to the baseline model. Note that in~\cite{anwar2015structured}, the accuracy of CIFAR10 using their method started to drop from $r=0.6$, which was much earlier than ours. Above experimental justify the effectiveness of applying IHT to train SDNN. 

\paragraph{Hard Thresholding vs Random Thresholding} We conduct experiments to justify the hard thresholding by replacing it with random thresholding in our method while keeping the other configurations unchanged. Specifically, we set $r=0.5$ in our experiments and replace Step 4 in Algorithm 1 with random thresholding in which half of parameters in each layer are randomly set to zero. However, we observe that deep models diverge quickly in Step 5 after random thresholding. The reason is that the random thresholding places deep models in bad conditions by truncating those important parameters.

\subsection{Results}
\label{sec:results}

\paragraph{CIFAR10} The CIFAR-10 dataset consists of 60,000 color images of 32
$\times$ 32 pixels in 10 classes. The total dataset is split into 50,000
training images and 10,000 testing images. Table~\ref{table:cifar10} compares the performance and \# parameters of SDNN and other state-of-the-art methods either when data augmentation is used or not. SDNN with different sparsity ratios are denoted with SDNN-\#$\times $ where \# is the reciprocal of the sparsity ratio. It is observed that when data augmentation is not applied, SDNN with sparsity ratio $r=0.5$ achieves the best result among all methods, reducing the error rate (ER) of the baseline model NIN by 1.71\%. Compared with RCNN-96 and RCNN-128 which have model sizes of 0.67M and 1.19M, respectively, SDNN-2$\times$ with a much smaller model size (0.49M) outperforms them by 0.61\% and 0.28\%, respectively, demonstrating that SDNN is able to significantly improve the generalization capability of deep models. To further test SDNN's performance in deep models with a larger size, we evenly increase the parameter of each layer in the original NIN by two times, resulting in a model which is four times as large as the original NIN. We denote the enlarged NIN as $\mbox{NIN}_{2}$ and test SDNN with sparsity ratio 
$r=0.5$
 (denoted as $\mbox{SDNN}_{2}$-2$\times$)
 on it. Compared with $\mbox{NIN}_{2}$, $\mbox{SDNN}_{2}$-2$\times$ is able to reduce the ER by 0.77\% and 1.72\% when data augmentation is applied or not, respectively, again verifies SDNN's capability to reduce overfitting. Note compared with ResNet which gets the lowest ER against all other methods, our method is only with a 0.02\% higher ER, but our model is much faster in testing since ResNet with 1.7M parameters has 110 layers while ours only has 9 layers. 

We also test SDNN with larger sparsity ratios when data augmentation is used. When $r=0.9$, SDNN-10x is slightly worse than the baseline NIN model (0.28\% higher on ER) with a significantly reduced model size (0.1M). We further train the SDNN-20x which has a sparsity ratio $r=0.95$ with  model size $0.05M$ to achieve the ER of 10.53\%, which is 1.64\% higher than the baseline NIN. Considering NIN is a network consisting of all convolution layers, this is still a satisfiable result.

\begin{table}[h]
\footnotesize
{
\caption{Comparison with existing models on CIFAR10. ER represents ``error rates'' and DA represents ``data augmentation''. The subscript for NIN and SDNN represents the ratio of the parameters' number in each layer w.r.t. the model without subscript. \emph{E.g.}, NIN$_{2}$ has as two times as many layer-wise parameters compared to baseline NIN. The number in ``\#$\times$'' for SDNN represents the model compression ratio of that model. The number after RCNN represents the number of convolution feature maps in each layer. Best results are shown in bold.}
\label{table:cifar10}
}

\centering
\begin{tabular}{l c c c}
\hline
Model                   & No. of Param.(M)  & ER w/ DA(\%) & ER w/o DA(\%) \\
\hline
NIN~\cite{nin}          & 0.97              & 10.41        & 8.81 \\
DSN~\cite{dsn}          & 0.97              & 9.69         & 7.97 \\
RCNN-96~\cite{RCNN}     & 0.67              & 9.31         & 7.37   \\
RCNN-128~\cite{RCNN}    & 1.19              & 8.98         & 7.24  \\
RCNN-160~\cite{RCNN}    & 1.86              & 8.69         & 7.09   \\
FitNet~\cite{fitnet}    & $\sim$2.5         & -            & 8.39 \\
Highway~\cite{highway}  & 2.3               & -            & 7.54 \\
ResNet~\cite{residual}  & 0.27              & -            & 8.75 \\
ResNet~\cite{residual}  & 1.7               & -            & \bf 6.43 \\
NIN$_{1/2}$             & 0.49              & 10.50        & 9.20 \\
NIN$_{2}$   & 3.92       & 9.2 & 8.17 \\
\hline
SDNN-2$\times$ &0.49          &\bf 8.70& 6.91    \\
SDNN-10$\times$ & 0.1         & - & 9.09 \\
SDNN-20$\times$ & \bf{0.05}        & - & 10.53 \\
SDNN$_{2}$-2$\times$ &1.79          &\bf 8.43& \bf6.45    \\
\hline
\end{tabular}
\end{table}

\begin{table}[h]
\footnotesize
\caption{Comparison with existing models on CIFAR100. Note since NIN~\cite{nin} did not report the results with DA, we therefore refer to the reimplementation of NIN in~\cite{apl} when DA is used.}
\label{table:cifar100}
\centering
\begin{tabular}{l c c c}
\hline
Model & No. of Param.(M) & ER w/ DA(\%)  & ER w/o DA(\%) \\
\hline
Maxout~\cite{goodfellow2013maxout} & \textgreater 5 & 38.57        & -\\
Prob. maxout~\cite{probmaxout} & \textgreater 5 & 38.14   & - \\
NIN~\cite{nin}    & \textgreater 5 & 35.68        & - \\
NIN~\cite{apl} & 0.97           & 35.96        & 32.70 \\
DSN~\cite{dsn}    & 0.97           & 34.57        & -\\
dasNet~\cite{DBLP:journals/corr/StollengaMGS14}    & -               & 34.50        & -\\
RCNN-96~\cite{RCNN}   & 0.67           & 34.18   & -\\
RCNN-128~\cite{RCNN}  & 1.19           & 32.59   & -\\
RCNN-160~\cite{RCNN}  & 1.86           & 31.75   & -\\
APL~\cite{apl}       & 0.97           & 34.40   & 30.83 \\
SReLU~\cite{SReLU}     & 0.97           & 31.23   & 29.91 \\  
NIN$_{2}$ & 3.81           & 31.29   & 29.12 \\
\hline
SDNN-2$\times$        & \bf{0.49}          &\bf 30.50   & \bf 29.51 \\
SDNN$_{2}$-2$\times$   & 1.79          &\bf 29.13   & \bf 27.14 \\

\hline
\end{tabular}
\end{table}

\paragraph{CIFAR100} CIFAR100 has 50,000/10,000 training/testing color images with resolution of $32\times 32$ pixels. Since the number of training images of each class in
CIFAR100 is only one tenth of that in CIFAR10, deep models trained on this dataset are prone to overfitting. Table \ref{table:cifar100} summaries the state-of-the-art methods on this dataset. It is obviously observed that either when the data augmentation is used or not, SDNN-2$\times$ achieves the lowest ER among all methods with the smallest model size. Specifically, compared with NIN model, SDNN-2$\times$ is able to reduce the ER by 5.18\% and 3.19\% either when data augmentation is used or not, respectively, which again demonstrates the advantages of SDNN. Moreover, compared with RCNN-96 which has a model size of 0.67M and achieves 34.18\% ER without using data augmentation, our method outperforms it significantly by reducing the ER by 3.68\% while simultaneously attaining a smaller model size (0.49M versus 0.67M). Like on CIFAR10, we also test SDNN$_{2}$-2$\times$ by adopting half of the model size of NIN$_{2}$. As can be seen from Table \ref{table:cifar100}, our method is able to further enhance the generalization capability of NIN$_{2}$ by reducing the ER significantly by 2.16\% /1.98\% when trained with / without data augmentation, respectively. 

\paragraph{MNIST}  MNIST is one of the standard datasets in the machine
learning community. It consists of 70,000 handwritten digits of 0 to 9 with 28$\times$28 resolution of pixels in gray scale format, which are split into 60,000/10,000 training/testing set. Table \ref{table:mnist} shows the comparison results, from which we can see SDNN-2$\times$ is able to achieve the best performance using the least parameters. Compared with the baseline model NIN, SDNN-2$\times$ greatly reduces the error rates from 0.47\% to 0.19\% on this heavily benchmarked dataset with only half of the NIN model size. To our best knowledge, this is the best performance ever achieved by methods without using ensembles and other preprocessing methods, which could further boost the performance of ours. 

\begin{table}

\begin{minipage}[l]{0.49\textwidth}%
\footnotesize
	\footnotesize \renewcommand{\arraystretch}{1}
	\caption{Error rates on MNIST without data augmentation. }
	\label{table:mnist} \centering
	\begin{tabular}{l c c} 
	\hline Model & \# Param.(M) & ER(\%) \\
	\hline
	Stocha. Pooling~\cite{stochasticpool}  & -    & 0.47 \\
	Maxout~\cite{goodfellow2013maxout}     & 0.42 & 0.45 \\ 
	NIN~\cite{nin}                         & 0.35 & 0.47 \\
	DSN~\cite{dsn}                         & 0.35 & 0.39 \\ 
	SReLU~\cite{SReLU}                     & 0.35 & 0.35 \\ 
	RCNN-96~\cite{RCNN}                    & 0.67 & 0.31 \\\hline
	SDNN-2$\times$                         & 0.18 & \textbf{0.19} \\
	\hline
	\end{tabular} \vspace{-0.3cm}
\end{minipage}\ \ \ \ \ \
	\begin{minipage}[r]{0.49\textwidth}%
	\footnotesize
	\caption{Comparison with existing models on ImageNet.}
	\label{table:imagenet}
	\centering
	\begin{tabular}{c c c}
	\hline
	Model & \# Param.(M) & Top-5 ER(\%) \\
	\hline
	Caffe Version\footnotemark            & 60  & 19.56   \\
	Han \& Pool~\cite{han2015learning}    & 6.7 & 19.67  \\
	Mem. Bound~\cite{collins2014memory}   & 15  & 19.6   \\
	\hline
	SDNN-2$\times$                        & 30  & \textbf{17.90}   \\
	SDNN-4$\times$                        & 15  & 18.75 \\
	\hline
	\end{tabular}
	\end{minipage}
\end{table}
\footnotetext{$\textrm{https://}\textrm{github}.\textrm{com/}\textrm{BVLC/}\textrm{caffe/}\textrm{tree/}\textrm{master/}\textrm{models/}\textrm{bvlc}_{\textrm{--}}\textrm{alexnet}$}

\paragraph{ImageNet}
To test the scalability of SDNN to large-scale datasets and model with bigger sizes, we conduct experiments on a much more challenging image classification task on 1000-class ImageNet dataset. As a renowned dataset in the vision research community, ImageNet contains about 1.2M training images, 50,000 validation images and 10,0000 testing images. To compare with other network pruning methods~\cite{han2015learning,collins2014memory}, we also use AlexNet as our baseline model according to the publicly available configurations in Caffe$^{1}$. Table \ref{table:imagenet} lists the performance of SDNN and other methods on ImageNet using AlexNet. Compared with the baseline model, our model with a two and four times compression ratio can further reduce the top-5 error rates by 1.66\% and 0.81\%, respectively, demonstrating the efficacy of SDNN on large-scale datasets. Compared with~\cite{han2015learning,collins2014memory}, both report higher error rates when pruning the deep model. SDNN is superior by boosting the performance  of deep models while largely reducing the model size at the same time. For example, compared with~\cite{collins2014memory}, SDNN-4$\times$ reduces its error rates significantly by 0.85\% while obtaining models with the same size.

\section{Conclusion}
In this paper, we proposed an iterative hard thresholding method (IHT) to improve the performance of the deep
neural network and reduce the size of parameters simultaneously. The training of SDNN using IHT consists of two alternative phases, \emph{i.e.} firstly performing hard thresholding to set connections with small magnitudes to zero and fine-tune the significant filters, and secondly, re-activating the freezing. Experiments conducted on four scale-various datasets, \emph{i.e.} CIFAR10, CIFAR100, MNIST and ImageNet using deep networks with different complexities, \emph{i.e.} NIN and AlexNet demonstrated that our method is able to significantly boost the discriminative capability of deep models while largely reducing their sizes simultaneously. 
 \bibliographystyle{plain} 
\bibliography{mybib}

\begin{thebibliography}{10}

\bibitem{apl}
Forest Agostinelli, Matthew Hoffman, Peter~J. Sadowski, and Pierre Baldi.
\newblock Learning activation functions to improve deep neural networks.
\newblock {\em CoRR}, abs/1412.6830, 2014.

\bibitem{anwar2015structured}
Sajid Anwar, Kyuyeon Hwang, and Wonyong Sung.
\newblock Structured pruning of deep convolutional neural networks.
\newblock {\em arXiv preprint arXiv:1512.08571}, 2015.

\bibitem{collins2014memory}
Maxwell~D Collins and Pushmeet Kohli.
\newblock Memory bounded deep convolutional networks.
\newblock {\em arXiv preprint arXiv:1412.1442}, 2014.

\bibitem{imagenet}
Jia Deng, Wei Dong, Richard Socher, Li-Jia Li, Kai Li, and Li~Fei-Fei.
\newblock Imagenet: A large-scale hierarchical image database.
\newblock In {\em CVPR}, 2009.

\bibitem{denton2014exploiting}
Emily~L Denton, Wojciech Zaremba, Joan Bruna, Yann LeCun, and Rob Fergus.
\newblock Exploiting linear structure within convolutional networks for
  efficient evaluation.
\newblock In {\em Advances in Neural Information Processing Systems}, pages
  1269--1277, 2014.

\bibitem{gong2014compressing}
Yunchao Gong, Liu Liu, Ming Yang, and Lubomir Bourdev.
\newblock Compressing deep convolutional networks using vector quantization.
\newblock {\em arXiv preprint arXiv:1412.6115}, 2014.

\bibitem{goodfellow2013maxout}
Ian~J Goodfellow, David Warde-Farley, Mehdi Mirza, Aaron Courville, and Yoshua
  Bengio.
\newblock Maxout networks.
\newblock {\em arXiv preprint arXiv:1302.4389}, 2013.

\bibitem{han2016}
Song Han, Huizi Mao, and William~J. Dally.
\newblock Deep compression: Compressing deep neural network with pruning,
  trained quantization and huffman coding.
\newblock {\em CoRR}, abs/1510.00149, 2015.

\bibitem{han2015learning}
Song Han, Jeff Pool, John Tran, and William Dally.
\newblock Learning both weights and connections for efficient neural network.
\newblock In {\em Advances in Neural Information Processing Systems}, pages
  1135--1143, 2015.

\bibitem{hassibi1993second}
Babak Hassibi and David~G Stork.
\newblock {\em Second order derivatives for network pruning: Optimal brain
  surgeon}.
\newblock Morgan Kaufmann, 1993.

\bibitem{residual}
Kaiming He, Xiangyu Zhang, Shaoqing Ren, and Jian Sun.
\newblock Deep residual learning for image recognition.
\newblock {\em arXiv preprint arXiv:1512.03385}, 2015.

\bibitem{SReLU}
Xiaojie Jin, Chunyan Xu, Jiashi Feng, Yunchao Wei, Junjun Xiong, and Shuicheng
  Yan.
\newblock Deep learning with s-shaped rectified linear activation units.
\newblock {\em CoRR}, abs/1512.07030, 2015.

\bibitem{kim2015compression}
Yong-Deok Kim, Eunhyeok Park, Sungjoo Yoo, Taelim Choi, Lu~Yang, and Dongjun
  Shin.
\newblock Compression of deep convolutional neural networks for fast and low
  power mobile applications.
\newblock {\em arXiv preprint arXiv:1511.06530}, 2015.

\bibitem{krizhevsky2009learning}
Alex Krizhevsky and Geoffrey Hinton.
\newblock Learning multiple layers of features from tiny images, 2009.

\bibitem{krizhevsky2012imagenet}
Alex Krizhevsky, Ilya Sutskever, and Geoffrey~E Hinton.
\newblock Imagenet classification with deep convolutional neural networks.
\newblock In {\em NIPS}, 2012.

\bibitem{lebedev2015fast}
Vadim Lebedev and Victor Lempitsky.
\newblock Fast convnets using group-wise brain damage.
\newblock {\em arXiv preprint arXiv:1506.02515}, 2015.

\bibitem{lecun1998gradient}
Yann LeCun, L{\'e}on Bottou, Yoshua Bengio, and Patrick Haffner.
\newblock Gradient-based learning applied to document recognition.
\newblock {\em Proceedings of the IEEE}, 86(11):2278--2324, 1998.

\bibitem{lecun1989optimal}
Yann LeCun, John~S Denker, Sara~A Solla, Richard~E Howard, and Lawrence~D
  Jackel.
\newblock Optimal brain damage.
\newblock In {\em NIPs}, volume~89, 1989.

\bibitem{dsn}
Chen-Yu Lee, Saining Xie, Patrick Gallagher, Zhengyou Zhang, and Zhuowen Tu.
\newblock Deeply-supervised nets.
\newblock {\em arXiv preprint arXiv:1409.5185}, 2014.

\bibitem{RCNN}
Ming Liang, Xiaolin Hu, and Bo~Zhang.
\newblock Convolutional neural networks with intra-layer recurrent connections
  for scene labeling.
\newblock In {\em NIPS}, 2015.

\bibitem{nin}
Min Lin, Qiang Chen, and Shuicheng Yan.
\newblock Network in network.
\newblock {\em arXiv preprint arXiv:1312.4400}, 2013.

\bibitem{fitnet}
Adriana Romero, Nicolas Ballas, Samira~Ebrahimi Kahou, Antoine Chassang, Carlo
  Gatta, and Yoshua Bengio.
\newblock Fitnets: Hints for thin deep nets.
\newblock {\em arXiv preprint arXiv:1412.6550}, 2014.

\bibitem{vgg}
Karen Simonyan and Andrew Zisserman.
\newblock Very deep convolutional networks for large-scale image recognition.
\newblock {\em arXiv preprint arXiv:1409.1556}, 2014.

\bibitem{probmaxout}
Jost~Tobias Springenberg and Martin Riedmiller.
\newblock Improving deep neural networks with probabilistic maxout units.
\newblock {\em arXiv preprint arXiv:1312.6116}, 2013.

\bibitem{dropout}
Nitish Srivastava, Geoffrey Hinton, Alex Krizhevsky, Ilya Sutskever, and Ruslan
  Salakhutdinov.
\newblock Dropout: A simple way to prevent neural networks from overfitting.
\newblock {\em The Journal of Machine Learning Research}, 15(1):1929--1958,
  2014.

\bibitem{highway}
Rupesh~Kumar Srivastava, Klaus Greff, and J{\"{u}}rgen Schmidhuber.
\newblock Highway networks.
\newblock {\em CoRR}, abs/1505.00387, 2015.

\bibitem{DBLP:journals/corr/StollengaMGS14}
Marijn~F. Stollenga, Jonathan Masci, Faustino~J. Gomez, and J{\"{u}}rgen
  Schmidhuber.
\newblock Deep networks with internal selective attention through feedback
  connections.
\newblock {\em CoRR}, abs/1407.3068, 2014.

\bibitem{googlenet}
Christian Szegedy, Wei Liu, Yangqing Jia, Pierre Sermanet, Scott Reed, Dragomir
  Anguelov, Dumitru Erhan, Vincent Vanhoucke, and Andrew Rabinovich.
\newblock Going deeper with convolutions.
\newblock {\em arXiv preprint arXiv:1409.4842}, 2014.

\bibitem{yosinski2014transferable}
Jason Yosinski, Jeff Clune, Yoshua Bengio, and Hod Lipson.
\newblock How transferable are features in deep neural networks?
\newblock In {\em Advances in Neural Information Processing Systems}, pages
  3320--3328, 2014.

\bibitem{yuan2013gradient}
Xiao-Tong Yuan, Ping Li, and Tong Zhang.
\newblock Gradient hard thresholding pursuit for sparsity-constrained
  optimization.
\newblock {\em arXiv preprint arXiv:1311.5750}, 2013.

\bibitem{stochasticpool}
Matthew~D. Zeiler and Rob Fergus.
\newblock Stochastic pooling for regularization of deep convolutional neural
  networks.
\newblock {\em CoRR}, abs/1301.3557, 2013.

\end{thebibliography}
\end{document}